\documentclass[journal]{IEEEtran}
\usepackage{amsmath,amsfonts}
\usepackage{algorithmic}
\usepackage{algorithm}
\usepackage{array}
\usepackage[caption=false,font=normalsize,labelfont=sf,textfont=sf]{subfig}
\usepackage{textcomp}
\usepackage{stfloats}
\usepackage{url}
\usepackage{verbatim}
\usepackage{graphicx}
\usepackage{cite}
\usepackage{makecell}
\usepackage{multirow}
\usepackage{tabularx}
\usepackage{amsmath}
\usepackage{xcolor}
\usepackage{hyperref} 
\usepackage{newtxtext}
\hypersetup{hidelinks}
\begin{document}
\title{MUST: Multi-Scale Structural-Temporal Link Prediction Model for UAV Ad Hoc Networks}

\author{Cunlai Pu, Fangrui Wu, Rajput Ramiz Sharafat, Guangzhao Dai, and Xiangbo Shu,~\IEEEmembership{Senior Member,~IEEE}
\thanks{Manuscript received April 19, 2021; revised August 16, 2021. This work was supported in part by the National Natural Science Foundation of China under Grant 62427808. \textit{(Corresponding author: Cunlai Pu.)}}
\thanks{Cunlai Pu, Fangrui Wu, Guangzhao Dai and Xiangbo Shu are with the School of Computer Science and Engineering, Nanjing University of Science and Technology, Nanjing 210094, China (e-mail: pucunlai, wufangrui, guangzhaodai, shuxb@njust.edu.cn)}
\thanks{Rajput Ramiz Sharafat is with the School of Computer Science and Technology,   University of Science and Technology of China, Hefei, 230026, China (e-mail: ramizrao@mail.ustc.edu.cn).}
}

\markboth{Journal of \LaTeX\ Class Files,~Vol.~14, No.~8, August~2021}%
{Shell \MakeLowercase{\textit{et al.}}: A Sample Article Using IEEEtran.cls for IEEE Journals}


\maketitle

\begin{abstract}
Link prediction in unmanned aerial vehicle (UAV) ad hoc networks (UANETs) aims to predict the potential formation of future links between UAVs. In adversarial environments where the route information of UAVs is unavailable, predicting future links must rely solely on the observed historical topological information of UANETs. However, the highly dynamic and sparse nature of UANET topologies presents substantial challenges in effectively capturing meaningful structural and temporal patterns for accurate link prediction. Most existing link prediction methods focus on temporal dynamics at a single structural scale while neglecting the effects of sparsity, resulting in insufficient information capture and limited applicability to UANETs. In this paper, we propose a multi-scale structural-temporal link prediction model (MUST) for UANETs. Specifically, we first employ graph attention networks (GATs) to capture structural features at multiple levels, including the individual UAV level, the UAV community level, and the overall network level. Then, we use long short-term memory (LSTM) networks to learn the temporal dynamics of these multi-scale structural features. Additionally, we address the impact of sparsity by introducing a sophisticated loss function during model optimization. We validate the performance of MUST using several UANET datasets generated through simulations. Extensive experimental results demonstrate that MUST achieves state-of-the-art link prediction performance in highly dynamic and sparse UANETs.
\end{abstract}

\begin{IEEEkeywords}
UANET,  link prediction, multi-scale feature, GAT, LSTM.
\end{IEEEkeywords}


\section{Introduction}
\IEEEPARstart{M}{ultiple} unmanned aerial vehicles (UAVs) can establish communication with each other in an ad hoc manner via wireless links, forming a UAV ad hoc network (UANET) \cite{pasandideh2023systematic,qiu2023integrated}. Leveraging their cooperative nature and high mobility, UAVs are widely used in critical applications, particularly military operations \cite{wang2024survey}. In adversarial engagements involving offensive and defensive maneuvers, accurately predicting potential future links within an adversary's UANET is essential for effective decision-making. However, in non-cooperative environments, key internal information, such as predefined mobility patterns and flight trajectories of adversarial UAVs, is typically inaccessible. Therefore, the link prediction task must rely exclusively on observed historical topological data, utilizing its structural and temporal information for accurate predictions.

Nevertheless, capturing meaningful information from topological data for link prediction in UANETs is a challenging task due to their highly dynamic nature and inherent sparsity~\cite{pasandideh2022review}.  The rapid flight speeds of UAVs during task execution, compared to traditional ground-based mobile nodes, cause frequent changes in network topology. When UAVs move beyond each other’s communication range, the link disconnects, and when they re-enter range, the link may be re-established.  This constant fluctuation is further influenced by weather conditions and obstacles such as buildings, which affect signal propagation and link quality. 
Additionally, the large areas over which UAVs are deployed result in significant distances between nodes, leading to a sparse network where only a limited number of nodes fall within each UAV's communication range, hindering link prediction.


Numerous link prediction methods have been proposed in the literature \cite{lu2011link,martinez2016survey,arrar2024comprehensive}, with many, such as common neighbor~\cite{liben2003link} and nonnegative matrix factorization  \cite{cai2010graph}, showing strong performance in static networks. However, these approaches overlook the temporal dynamics of networks, making them unsuitable for dynamic environments like UANETs. To address this, researchers have tried incorporating temporal information into traditional methods. Some have enhanced static common neighbor-based techniques with dynamic data \cite{yao2016link}, while others have stacked topological features from multiple historical snapshots \cite{he2024sequential} or applied matrix factorization to auxiliary snapshots based on historical data \cite{gao2011temporal}. Despite these efforts, the performance of these methods remains less than ideal and continues to require further refinement.

Advancements in deep learning have led to the use of graph neural networks (GNNs) for capturing structural information and recurrent neural networks (RNNs) for modeling temporal dependencies in dynamic network link prediction tasks \cite{qin2023temporal}. These methods have shown promising results across various applications. However, they tend to focus on a single structural scale, limiting their predictive accuracy. In UANETs, topology spans multiple scales—from individual UAVs at the micro scale to UAV communities at the meso scale, and the entire network at the macro scale. To improve prediction accuracy, it is crucial to simultaneously consider structural information across all these scales.


To the best of our knowledge, no link prediction methods have been specifically tailored for UANETs. As a result, the performance of existing dynamic link prediction techniques may degrade when applied to UANETs, given their unique characteristics.
In this paper, we investigate the problem of UANET link prediction for the first time and propose a multi-scale structural-temporal link prediction model, named MUST. 
In this model, we analyze structural features across multiple scales: individual UAVs at the micro level, UAV communities at the meso level, and entire network at the macro level. To achieve this, we use a variant of graph attention networks (GATs) \cite{sankar2020dysat} to extract micro-scale structural features from each network snapshot. These features are then processed using meso pooling and macro pooling to generate meso- and macro-scale structural representations, resulting in a comprehensive multi-scale representation for each snapshot. To capture the temporal evolution of these multi-scale features, we employ long short-term memory (LSTM) networks \cite{van2020review} across consecutive network snapshots. Additionally, to address network sparsity, we design a sophisticated loss function that prioritizes the contributions of existing links during training, significantly improving the model's predictive performance. Our main contributions are as follows:
\begin{enumerate}
	\item 	We formalize the link prediction problem in UANETs and present a novel model designed to address the challenges of high dynamics and topological sparsity unique to these networks.  
	\item Our model extracts topological features at multiple levels, constructing multi-scale structural representations. By capturing the temporal evolution of these multi-scale features, we generate comprehensive structural-temporal representations for UANET link prediction.
	\item To evaluate our model, we simulate various UAV mobility patterns and create extensive UANET datasets. Experimental results demonstrate that our model outperforms existing methods, achieving state-of-the-art prediction accuracy.
\end{enumerate}

The remainder of this paper is structured as follows: Section \hyperref[RelatedWork]{\uppercase\expandafter{\romannumeral2}} reviews related work on link prediction, highlighting existing approaches and their limitations. Section \hyperref[ProblemDefinition]{\uppercase\expandafter{\romannumeral3}} formally defines the UANET link prediction problem. In Section \hyperref[Methodology]{\uppercase\expandafter{\romannumeral4}}, we provide a detailed explanation of our proposed link prediction model. Section \hyperref[Experiment]{\uppercase\expandafter{\romannumeral5}}  presents and analyzes the experimental results. Finally, Section \hyperref[Conclusion]{\uppercase\expandafter{\romannumeral6}} concludes the paper and outlines potential directions for future research.

\section{Related Work}\label{RelatedWork}
Existing link prediction methods for both static and dynamic networks can be broadly classified into three categories. The first category includes methods based on traditional network topological properties. These methods are intuitive and computationally efficient but are limited to capturing local network features, which restricts their performance. The second category includes methods based on matrix factorization, which capture global and latent network features. However, these methods are computationally expensive, making them impractical for large-scale networks, and their performance still leaves room for improvement. The third category includes methods based on neural networks and deep learning, which capture highly nonlinear relationships and enable end-to-end optimization based on training data, offering powerful learning capabilities.

\vspace{-10pt}
\subsection{Methods Based on Traditional Topological Features}
Traditional topological features have long been utilized to measure the similarity between pairs of nodes, with the assumption that higher similarity indicates a greater likelihood of a link between them. Common examples of such features include Common Neighbour (CN) \cite{liben2003link}, Adamic-Adar \cite{adamic2003friends}, and Resource Allocation \cite{zhou2009predicting}. These methods perform adequately in static networks that evolve predictably based on these similarity metrics. However, their effectiveness diminishes in complex and dynamic networks. To address this limitation, Yao et al. \cite{yao2016link} proposed a dynamic link prediction method based on CN, integrating the network's dynamic characteristics to improve accuracy. Similarly, He et al. \cite{he2024sequential} introduced the Top-Sequential-Stacking algorithm, which combines static topological features from sequential time-layer networks into feature vectors, enabling better link prediction in dynamic settings. Although these enhancements improve prediction performance, traditional methods often fall short when compared to advanced techniques designed to handle the complexities of modern dynamic networks.

\vspace{-10pt}
\subsection{Methods Based on Matrix Factorization}
Matrix factorization techniques are powerful tools for uncovering the structural features of a graph by performing low-rank decomposition on graph-related matrices, such as the adjacency matrix.  Cai et al. \cite{cai2010graph} pioneered the use of a graph-regularized non-negative matrix factorization method for link prediction, specifically designed for static graphs. However, static methods fall short in dynamic networks where links evolve over time. To tackle this, Gao et al. \cite{gao2011temporal} introduced an approach that integrates historical graph snapshots into auxiliary snapshots during matrix factorization, enabling the identification of evolving link patterns. Despite its innovation, this method overlooks the interconnections between snapshots, limiting its predictive power. To address this, Ma et al. \cite{ma2018graph} developed a novel matrix factorization algorithm that leverages historical networks as a regularizer, effectively preserving relationships among snapshots. Building on this, Yu et al. \cite{yu2017link} proposed an advanced method that models network representation as a function of time, ensuring consistency between spatial and temporal factors for enhanced dynamic link prediction. These advancements highlight the growing sophistication of matrix factorization techniques in capturing the complexities of dynamic networks.

\vspace{-10pt}
\subsection{Methods Based on Neural Networks and Deep Learning}
Recent advancements in link prediction have seen an increasing reliance on neural networks and deep learning techniques. Drawing inspiration from the skip-gram model \cite{preethi2020word} in natural language processing, methods like DeepWalk \cite{perozzi2014deepwalk} and Node2Vec \cite{grover2016node2vec} generate random walk paths, which are then converted into low-dimensional node embeddings through training. These embeddings serve as the foundation for link prediction tasks. While random walk-based approaches are effective, GNNs have proven to be more powerful in capturing structural information. GNNs propagate node information to neighboring nodes and aggregate it, enabling a richer understanding of the graph's structure. Graph Attention Networks (GATs) \cite{velickovic2017graph}, a specialized type of GNN,  introduce attention mechanisms that dynamically adjust the importance of nodes, improving performance, especially in graphs with varying node relevance. Similarly, Graph Convolutional Networks (GCNs) \cite{kipf2016semi} use spectral filters to propagate local graph information, capturing local features effectively. In contrast, GraphSAGE \cite{hamilton2017inductive} introduces a scalable approach by employing sampling techniques to aggregate features from a node's neighbors, making it well-suited for large-scale graphs.  While these methods excel in static graph scenarios, they require additional processing and adaptation for effective use in dynamic network link prediction.

Embedding-based methods \cite{cui2018survey} have significantly advanced the analysis of dynamic networks. Goyal et al. \cite{goyal2020dyngraph2vec} introduced Dyngraph2vec, a network embedding model designed to capture temporal patterns, demonstrating that a deeper understanding of network dynamics enhances dynamic link prediction performance. Building on this, Sankar et al. \cite{sankar2020dysat} proposed DySAT, a dynamic network embedding algorithm that leverages structural self-attention layers to extract information from individual snapshots and temporal self-attention layers to model the graph's evolution over time. Zhang et al. \cite{zhang2023attentional} developed AMCNet, which focuses on learning intrinsic correlations across different structural scales of network evolution. By using high-scale representations to guide the learning of low-scale representations, AMCNet generates robust multi-scale representations for dynamic link prediction. Despite their strengths, these embedding-based methods share a common limitation: they lack an end-to-end training framework, as embedding learning and link prediction are conducted separately, which may hinder their overall performance.

Advancing the field of end-to-end dynamic link prediction, researchers have developed innovative models to tackle the evolving nature of networks. Pareja et al. \cite{pareja2020evolvegcn} introduced EvolveGCN, which dynamically updates GCN parameters using RNNs, enabling the model to adapt to temporal changes. Similarly, Chen et al. \cite{chen2019lstm} proposed E-LSTM-D, an encoder-decoder-based model that employs stacked Long Short-Term Memory (LSTM) networks to capture historical trends and effectively model network evolution. Qin et al. \cite{qin2023high} integrated generative adversarial networks with multiple optimization objectives to deal with network dynamics, enabling the generation of high-quality link prediction results. Expanding the scope, Pham et al.  \cite{pham2021comgcn}  proposed ComGCN, a community-driven model that utilizes Node2Vec for generating initial node features and leverages GCNs to create multi-scale features for dynamic link prediction. However,  ComGCN lacks the integration of link weight information, which is crucial for capturing link quality, particularly in applications like UAV networks where accurate link prediction heavily depends on this metric.

The link prediction methods discussed earlier are generally not tailored for UANETs and have yet to be evaluated in this context. Static link prediction methods lack direct applicability to the dynamic nature of UANETs. Although previous dynamic methods show promise for adaptation, their effectiveness remains uncertain, as many rely solely on single-scale features, disregard historical communication link quality, or struggle to address network sparsity. 
In contrast, our proposed model overcomes these limitations by integrating multi-scale features and effectively managing the high dynamics and sparsity characteristic of UANETs. This comprehensive approach enables more accurate predictions of future links, addressing the unique challenges of UANET environments.

\section{Problem Definition}\label{ProblemDefinition}
In a UANET, multiple UAVs operate autonomously according to predefined rules to accomplish specific missions. They can establish ad hoc communication links with each other when within the maximum communication range, which is influenced by factors such as transmitter power, receiver sensitivity, and environmental conditions. Despite this capability, the quality of these links often fluctuates due to challenges like positional shifts, obstacles, path loss, interference, and adverse weather conditions.
 In our work, we assign a weight to each link to quantify its reliability. Specifically, the weight of a link between UAV nodes $i$ and $j$ is defined as follows:
 
\vspace{-10pt}
\begin{equation}
\label{1}
    w_{ij}=\begin{cases}
\dfrac{r-d(i,j)}{r} \log_2\left(1+S(i,j)\right), & \text{if } d(i,j) \leq r, \\
0, & \text{otherwise}.
    \end{cases}
\end{equation}
where $r$ represents the maximum communication range, $d(i, j)$ denotes the distance between UAVs $i$ and $j$, and $S(i, j)$ indicates  the signal-to-noise ratio (SNR) of the communication channel between them, which follows a normal distribution in our work.  Generally, a shorter distance and a higher SNR  indicate a more stable link. Note that if the link weight is zero, the connection between the UAVs is considered absent.

By using Eq. \eqref{1}, we derive all the links  with their corresponding weights, resulting in a weighted graph representation for each UANET snapshot, denoted as $\{G_1,G_2,\ldots\}$. Here, $G_t = (V, L_t, \Omega_t)$ represents the snapshot at time $t$, where $V$ is the set of nodes, $L_t$ is the set of links, and $\Omega_t$ is the set of link weights at time $t$. The weighted adjacency matrix of $G_t$ is denoted as $M^t \in \mathbb{R}^{|V| \times |V|}$, where each element $M^t_{ij}$ represents the weight of the link between nodes $i$ and $j$ at time $t$.
Each row of $M^t$ reflects the connectivity of the corresponding node to all other nodes. Therefore, we include each row of this matrix as part of the node attribute. To further differentiate between UAVs, we also incorporate the 32-bit IP address of each UAV as part of its node attribute. Let $X^t \in \mathbb{R}^{|V| \times D_X}$  represent the overall node attribute matrix at time $t$, where $D_X$  denotes the dimensionality of node attributes. We define $X^t = [X_{\text{ip}} || M^t]$,  where $[\cdot || \cdot]$ denotes the concatenation operation, and $X_{\text{ip}}\in \mathbb{R} ^{|V|\times 32}$ is a matrix containing the IP addresses, with each row corresponding to the IP address of a node. Due to the concatenation, we have $D_X = |V| + 32$.

\begin{figure}[!t]
\vspace{-10pt}
\centering
\includegraphics[width=1\linewidth]{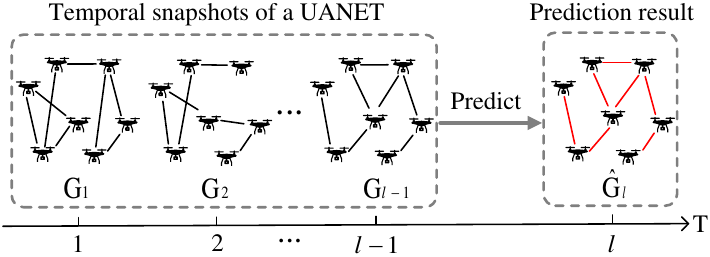}
\caption{An illustration of the link prediction problem in UANETs, where $G_1, G_2, \ldots, G_{l-1}$ represent the historical topology snapshots, and the links in $G_l$ are the target to be predicted. }
\label{fig1}
\vspace{-10pt}
\end{figure}
As shown in Fig. \ref{fig1}, given a sequence of UANET topology snapshots of length $l-1$, i.e., $\{G_1, G_2, \ldots, G_{l-1}\}$, we aim to predict the links in $G_l$. The unweighted adjacency matrix $A^t \in \mathbb{R}^{|V| \times |V|}$ is used to indicate the existence of links at time $t$, where the element $A^t_{ij} = 1$ if a link exists between nodes $i$ and $j$ at time $t$, and $A^t_{ij} = 0$ otherwise. Thus, the task of dynamic link prediction in UANETs can be formally expressed as follows:
\begin{equation}
	\label{2}
	\hat{A}^l=\mathcal{F}\big( \{M^1,M^2,\ldots,M^{l-1}\} ,\{X^1,X^2,\ldots,X^{l-1}\}\big).
\end{equation}
The input to the function $\mathcal{F}$ consists of the given sequences of weighted adjacency matrices $\{M^1, M^2, \ldots, M^{l-1}\}$ and attribute matrices $\{X^1, X^2, \ldots, X^{l-1}\}$, while the output is the prediction of unweighted adjacency matrix $\hat{A}^l$. The  primary objective of UANET link prediction is to accurately learn this mapping function $\mathcal{F}$. 

\begin{figure*}[!t]
\vspace{-10pt}
\includegraphics[width=1\textwidth]{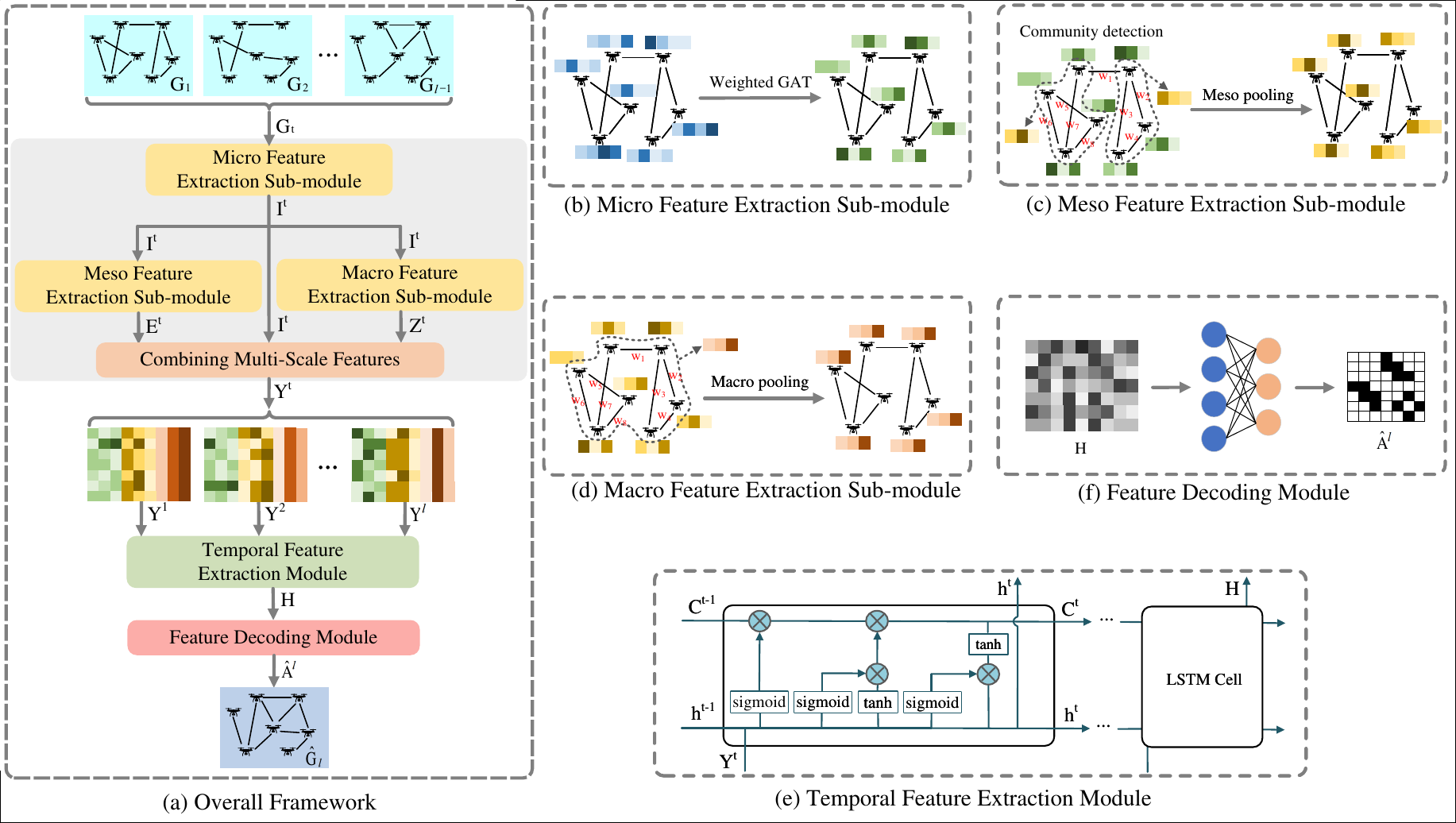}
\caption{(a) The overall framework of our proposed model, MUST, comprises three main components: a multi-scale feature extraction module, a temporal feature extraction module, and a feature decoding module. The multi-scale feature extraction module is further divided into (b) a micro-scale feature extraction sub-module, which employs a weighted GAT, (c) a meso-scale feature extraction sub-module, where meso-level pooling is applied, and (d) a macro-scale feature extraction sub-module, where macro-level pooling is performed. The temporal feature extraction module leverages (e) a  LSTM network to capture the dependencies within temporal topology sequences. Finally, the feature decoding module utilizes (f) a two-layer fully connected neural network to generate the prediction results.}
\label{fig2}
\vspace{-10pt}
\end{figure*}
\section{Methodology}\label{Methodology}
In this section, we first provide a detailed description of the proposed prediction model. Next, we introduce the model optimization process, focusing on the design of the loss function. Finally, we provide a comprehensive complexity analysis of the model.

\vspace{-10pt}
\subsection{Model Framework}
As illustrated in Fig. \ref{fig2}, our prediction model, MUST, comprises three components: the multi-scale feature extraction module, the temporal feature extraction module, and the feature decoding module. To predict links in a dynamic UANET, two types of information are critical: structural information and temporal information. 
The multi-scale feature extraction module captures network structural details at micro, meso, and macro levels, ensuring a comprehensive representation of the network. The temporal feature extraction module then models the complex temporal dependencies within these multi-scale structural features across various graph snapshots. Finally, the feature decoding module processes the output from the temporal feature extraction module to predict the adjacency matrix effectively. A detailed explanation of each module is provided below.

\subsubsection{Multi-Scale Feature Extraction Module}
The multi-scale feature extraction module is structured into three interconnected sub-modules: micro, meso, and macro feature extraction. Each sub-module plays a distinct role in capturing structural information at varying levels of granularity, ensuring a comprehensive understanding of the network's architecture.

\textit{Micro Feature Extraction Sub-module}.
The micro feature extraction sub-module takes two inputs: the weighted adjacency matrix $M^t$ and the attribute matrix $X^t$. Its output is the micro-level feature matrix $I^t \in \mathbb{R}^{|V| \times D}$, where $D$ denotes the dimensionality of the micro features. In the context of UANETs, these micro features are extracted at the UAV level, capturing fine-grained structural and attribute information critical for accurate predictions.

Specifically, this sub-module utilizes a GAT to extract the micro-level structural features of UANETs. GAT computes attention weights for each node and its first-order neighbors, aggregating information from neighboring nodes based on these weights \cite{velickovic2017graph}. Given that UANETs are weighted networks in our setting, we employ a weighted version of GAT \cite{sankar2020dysat} to account for link weights effectively. For each graph snapshot  $G_t$,  GAT  calculates  the attention weight  between nodes as follows:

\vspace{-10pt}
\begin{align}
e^t_{ij} &= \sigma \left( M^t_{ij}a^T\left[ W_GX^t_i || W_GX^t_j \right] \right), \label{3} \\
\alpha ^t_{ij} &= \frac{\exp \left( e^t_{ij} \right)}{\sum\limits_{m\in \mathcal N_i}{\exp \left( e^t_{im} \right)}}, \ i,j \in V. \label{4}
\end{align}
Eq. \eqref{3} computes the attention score $e^t_{ij}$ between node $i$ and each of its neighboring nodes $j$, Here, $M^t_{ij}$ denotes the entry in row $i$ and column $j$ of the weighted adjacency matrix $M^t$, representing the link weight between nodes $i$ and $j$. The vector $X^t_i \in \mathbb{R}^{D_X}$ represents the initial attribute of node $i$, while $W_G \in \mathbb{R}^{D \times D_X}$ and $a \in \mathbb{R}^{2D}$ are learnable parameters. The operator $[\cdot || \cdot]$ signifies concatenation, and $\sigma(\cdot)$ is a non-linear activation function, implemented as the $LeakyReLU$ function in this work.
Eq. \eqref{4} calculates the attention weight $\alpha^t_{ij}$, which reflects the structural importance of node $j$ to node $i$ at the micro level, where $\mathcal{N}_i$ is the set of first-order neighbors of node $i$. Finally, The micro-level structural feature for each node $i$ in graph snapshot $G_t$ is derived using the following equation:

\begin{equation}
\label{5}
I^t_i=\sigma \big( \sum_{j\in \mathcal N_i}{\alpha ^t_{ij}}W_GX^t_i\big),\ i\in V,
\end{equation}
where $\sigma(\cdot)$ represents a non-linear activation function,  specifically the $ELU$ function used in this work.

\textit{Meso Feature Extraction Sub-module.} The input to the meso feature extraction sub-module consists of the weighted adjacency matrix \(M^t\) and the micro feature matrix \(I^t\), while the output is the meso feature matrix of the nodes \(E^t \in \mathbb{R}^{|V| \times D}\). In UANETs, UAVs often work together to perform complex tasks, forming communities where communication within a community is more frequent than with UAVs from other communities. This leads to denser intra-community links compared to inter-community links, a pattern commonly seen in complex networks such as social and biological networks \cite{newman2018networks}. Since meso-structural information of UAV community effectively captures the evolutionary patterns of UANETs, we leverage this information to enhance structural prediction tasks.

We employ the Louvain community detection algorithm \cite{blondel2008fast} to identify communities in UANETs, which is widely used due to its computational efficiency and effectiveness. Following community division, the network $G_t$ is divided into communities \(\{G_{t,1}, G_{t,2}, \ldots, G_{t,c},\dots\}\), where each subgraph \(G_{t,c} = (V_{t,c}, E_{t,c})\)  represents the \(c\)-th community of \(G_t\). Here \(V_{t,c}\) and \(E_{t,c}\) denote the set of nodes and links within the community, respectively.
Inspired by \cite{zhang2023attentional}, information from lower structural scales can be transformed into higher ones  through pooling operations. Specifically, we propose a novel meso-scale pooling method that aggregates the micro features of nodes within community to derive the meso feature of this community: 
\vspace{-10pt}
\begin{equation}
\label{6}
E^t_c=\sum_{k\in V_{t,c}}{\omega _k I^t_k}\ ,\ \ \ \omega _k=\dfrac{\sum\limits_{m\in {\mathcal N_k,V_{t,c}} }{M^t_{km}}}{\sum\limits_{i,j\in V_{t,c}}{M^t_{ij}}},
\end{equation}
where \(\omega_k\) represents the weight coefficient corresponding to node \(k\), \(\mathcal{N}_k\) denotes the set of first-order neighbors of node \(k\), and \(E_{t,c} \in \mathbb{R}^D\) is the meso-scale feature extracted from the community \(G_{t,c}\). According to Eq. \eqref{6}, \(\omega_k\) is proportional to the total weight of the links associated with node $k$ within the community. This weighted pooling approach utilizes link weight to more effectively capture intra-community link quality among UAVs. Finally, by stacking the meso-scale features of all nodes, the meso-scale feature matrix \(E^t\) is constructed, with each row representing the meso feature of a node derived from its respective community.

\textit{Macro Feature Extraction Sub-module.} The macro feature extraction sub-module takes the weighted adjacency matrix $M^t$ and the micro-scale feature matrix $I^t$ as inputs and outputs the macro-scale feature matrix $Z^t \in \mathbb{R}^{|V| \times D}$. At this level, features are extracted directly from the entire UANET. The macro-scale pooling method aggregates the micro-scale features of all UAVs as follows:
\vspace{-5pt}
\begin{equation}
\label{7}
Z^t_s=\sum_{k\in V}{\phi _kI^t_k}\ ,\ \ \ \phi _k=\dfrac{\sum\limits_{m\in \mathcal N_k}{M^t_{km}}}{\sum\limits_{i,j\in V_t}{M^t_{ij}}},
\end{equation}
where $Z^t_s \in \mathbb{R}^D$ represents the macro feature extracted from the entire graph, and $\phi_k$ represents the weight coefficient associated with node $k$. This coefficient is calculated as the ratio of the sum of link weights connected to node $k$ to the total weight of all links in the graph. By stacking $Z^t_s \in \mathbb{R}^D$ for all nodes ($|V|$ times), the final macro-scale feature matrix $Z^t$ is obtained, where each row is identical to $Z^t_s$. Similar to meso-scale pooling, macro-scale pooling effectively utilizes the rich information contained in the link weights of UAV communication networks.

\textit{Combining Multi-Scale Features.} At this stage, the micro, meso, and macro features of the nodes are extracted. These features are then seamlessly combined as follows:

\begin{equation} \label{8} Y^t = [I^t || E^t || Z^t], \end{equation}
where $Y^t$ represents the multi-scale node feature matrix formed by directly concatenating the features from the three scales. This straightforward yet efficient approach ensures the effective utilization of information captured at each scale, offering a clear advantage over other feature-merging methods.

\subsubsection{Temporal Feature Extraction Module}
Through multi-scale feature extraction, we obtain the sequence of multi-scale node feature matrices $\{Y^1, Y^2, \ldots, Y^{l-1}\}$, corresponding to the sequence of graph snapshots $\{G_1, G_2, \ldots, G_{l-1}\}$. Given the high dynamics of UANETs, it is essential to capture the temporal dependencies between  node structural features across these snapshots  to accurately predict the links in $G_l$.

To capture these temporal dependencies, we utilize a LSTM network, which is well-suited for sequential data due to its capability to retain long-term dependencies. As illustrated in Fig. \ref{fig2}(e), the LSTM architecture comprises forget, input, and output gates that control  the flow and storage of information. We employ a stacked LSTM with $K$ layers to learn the temporal features. The computation of the forget gate is given by the following equation:
\begin{equation}
\label{9}
f^t=\sigma \left( W_f\left[ h^{t-1}||Y^t \right] +b_f \right), 
\end{equation}
where $\sigma(\cdot)$ denotes the $sigmoid$ activation function, $W_f$ and $b_f$ are the learnable weights and biases, and $h_{t-1}\in \mathbb{R} ^{|V|\times D_L}$ represents the output of the LSTM unit at the previous time step, with $D_L$ being the dimension of the LSTM hidden state. The computation of the input gate is given by:
\begin{equation}
\label{10}
\begin{split}
&i^t=\sigma \left( W_i\left[ h^{t-1}||Y^t \right] +b_i \right) ,
\\
&\tilde{C}^t=\sigma \left( W_C\left[ h^{t-1}||Y^t \right] +b_C \right).
\end{split}
\end{equation}
This calculation consists of two parts: the first part computes $i^t\in \mathbb{R} ^{|V|\times D_L}$, which controls the retention of new input information, where $\sigma(\cdot)$ is the $sigmoid$ activation function and $W_i$ and $b_i$ are the learnable weights and biases. The second part computes the candidate cell state $\tilde{C}^t\in \mathbb{R} ^{|V|\times D_L}$, where $\sigma(\cdot)$ is the $tanh$ activation function, and  $W_C$ and $b_C$ are the learnable weights and biases. The new cell state is derived from both the forget and input gates, as shown in the following equation:
\begin{equation}
\label{11}
C^t=f^t\odot C^{t-1}+i^t\odot \tilde{C}^t, 
\end{equation}
where $\odot$ represents the Hadamard product. The computation of the output gate is defined as:
\begin{equation}
\label{12}
\begin{split}
&o^t=\sigma \left( W_o\left[ h^{t-1}||Y^t \right] +b_o \right), 
\\
&h^t=o^t\odot \sigma \left( C^t \right). 
\end{split}
\end{equation}
The first part computes $o^t\in \mathbb{R} ^{|V|\times D_L}$, which controls the information to be output from the new cell state, while the second part computes the hidden state $h^t$, where $\sigma(\cdot)$ is the $tanh$ activation function.

By sequentially inputting $\{Y^1, Y^2, \dots, Y^{l-1}\}$ into the temporal feature extraction module, the final LSTM unit in the $K$-th layer produces the node temporal feature matrix $H \in \mathbb{R}^{|V| \times D_L}$.  This matrix captures the temporal dependencies of the nodes across all $l-1$ time steps and is used for link prediction at time step $l$.

\subsubsection{Feature Decoding Module}
The feature decoding module utilizes a neural network consisting of two fully connected  layers. The computations for these layers are as follows:
\begin{equation}
\label{13}
\hat{A}^t=\sigma \left( W_{d}^{\left( 2 \right)}\sigma \left( W_{d}^{\left( 1 \right)}H+b_{d}^{\left( 1 \right)} \right) +b_{d}^{\left( 2 \right)} \right),  
\end{equation}
where $H$ is the output from the temporal feature extraction module, and $W_{d}^{\left( 1 \right)}$, $b_{d}^{\left( 1 \right)}$, $W_{d}^{\left( 2 \right)}$ and $b_{d}^{\left( 2 \right)}$ are the learnable weights and biases. The activation function for the first layer is $ReLU$, while the second layer uses the $sigmoid$ function. Each element of the output matrix $\hat{A}^t$ represents the probability of the existence of a link between two nodes. To refine the prediction results, we employ the following commonly used operations: 
\vspace{-10pt}
\begin{equation}
\label{14}
\begin{split}
&\hat{A}^t_{ij} \longleftarrow 0\ ,\ \ i,j\in V,
\\
&\hat{A}^t \longleftarrow \frac{( \hat{A}^t+\hat{{A}^t}^T )}{2}.
\end{split}
\end{equation}
The first operation sets the diagonal elements to zero, ensuring no self-loops in the UANETs. The second operation enforces the symmetry of the adjacency matrix, reflecting the undirected nature of links in UANETs.

\vspace{-10pt}
\subsection{Model Optimization}\label{ModelOptimization}
Given a sequence of UAV graph snapshots $\{G_1, G_2, \ldots, G_N\}$, a sliding window of length $l<N$ is applied with a stride of one. At each step, a sub-sequence of length $l$ is extracted, resulting in $N-l+1$ sub-sequences. In each sub-sequence, the first $l-1$ snapshots $\{G_1, G_2, \ldots, G_{l-1}\}$ represent the known information, while the final snapshot $G_l$ serves as the prediction target. These $N-l+1$ samples are then split into training and test sets for experimental evaluation.
The loss function comprises reconstruction error and regularization terms. A detailed description of the loss function is provided below.

\textit{Reconstruction Error.} The predicted adjacency matrix $\hat{A}^t$ is expected to closely approximate the true adjacency matrix $A^t$. The squared $L_2$ distance is employed to measure the difference between these two matrices. However, the UANET topology is highly sparse, with significantly more zero elements than non-zero elements in $A^t$. To ensure the model focuses more on the non-zero elements, a parameter matrix $S$ is introduced to weight the $L_2$ distance. The reconstruction error is defined as:
\vspace{-10pt}
\begin{equation}
\label{15}
\mathcal{L}_{rec}=\sum_{i,j} S_{ij} (A^t_{ij} - \hat{A}^t_{ij})^2, \quad 
S_{ij}=\begin{cases}
	\varepsilon, & A^t_{ij} = 1, \\
	1, & A^t_{ij} = 0.
\end{cases}
\end{equation}
where $S_{ij}$ denotes the element at the $i$-th row and $j$-th column of matrix $S$, and $\varepsilon > 1$ is the weight assigned to non-zero elements. However, as noted in \cite{yin2022se}, simply using $L_2$ distance to calculate the reconstruction error may cause many elements in $\hat{A}^t$ to approach 0.5. In the corresponding $\hat{G}^t$, this results in many nodes being connected with lower probabilities, which contradicts the sparsity of UANETs. To encourage sparse predictions, we incorporate an $L_1$ regularization term. The final reconstruction error is expressed as:
\begin{equation}
\label{16}
\mathcal{L}_{sprec}=L_{rec}+\beta || \hat{A}^t || _1,
\end{equation}
where $\beta$ is a hyperparameter that  controls the strength of the regularization.


\textit{$L_2$ Regularization.} 
To prevent overfitting and improve the model's generalization ability, we apply $L_2$ regularization to all learnable parameters of the model. The regularization loss is defined as:
\begin{equation}
	\label{17}
	\mathcal{L}_{reg} = \sum_{i} ||\Theta_i||_F^2,
\end{equation}
where $\Theta_i$ represents the $i$-th parameter matrix of the model.

During  model training,  graph snapshots $\{G_1, \ldots, G_{l-1}\}$ are input sequentially, and the corresponding predictions $\{\hat{A}_2, \ldots, \hat{A}_l\}$ are generated in a similar fashion. Each prediction is associated with a corresponding loss. To emphasize recent snapshots over earlier ones, we introduce a decay factor, $\rho^t$, which adjusts the loss function by assigning higher weights to the losses of more recent predictions. Specifically, this is expressed as: 
\begin{equation}
\label{18}
\rho ^t=\eta^{l-t}\ ,\ t\in \left[ 2, l \right] ,
\end{equation}
where $\eta \in (0,1)$ is a hyperparameter. 
The total loss function for the model is then given by:
\begin{equation}
\label{19}
\mathcal{L}=\rho ^t \mathcal{L}_{sprec} +\lambda \mathcal{L}_{reg},
\end{equation}
where  $\lambda$ is a hyperparameter that controls the relative contribution of regularization term.

Through backpropagation, the gradients of the loss function with respect to the model parameters are computed. The Adam optimization algorithm \cite{reyad2023modified} is then used to iteratively update these parameters.  After a sufficient number of iterations, the model parameters converge to their optimal values. The overall training procedure of MUST is described in Algorithm \ref{alg:1}.

\begin{algorithm}[!t]
\caption{Training Procedure of MUST.}
\label{alg:1}
\begin{algorithmic}[1]
\REQUIRE Training snapshot sequence $\{G_1, \ldots, G_{N_0}\}$, number of training snapshots $N_0$, number of training epochs $T$, window size $l$.
\ENSURE Trained model.
\STATE Randomly initialize the model parameter $\Theta$.
\FOR{$epoch = 1$ to $T$}
    \FOR{$n = l$ to $N_0$}
    \STATE Extract the subsequence from $\{G_1, \ldots, G_{N_0}\}$ with indices ranging from $n - l + 1$ to $n - 1$.
        \FOR{$i = n - l + 1$ to $n - 1$}
            \STATE Input $G_i$ into micro feature extraction module.
            \STATE Compute micro feature $I_i$ using Eqs. \eqref{3}--\eqref{5}.
            \STATE Input $I_i$ into meso feature extraction module.
            \STATE Compute meso feature $E_i$ using Eq. \eqref{6}.
            \STATE Input $I_i$ into macro feature extraction module.
            \STATE Compute macro feature $Z_i$ using Eq. \eqref{7}.
            \STATE Compute multi-scale feature $Y_i$ using Eq. \eqref{8}.
            \STATE Input $Y_i$ into temporal feature extraction module.
            \STATE Compute temporal feature $H$ using Eqs. \eqref{9}--\eqref{12}.
            \vspace{-12pt}
            \STATE Input $H$ into feature decoding module.
            \STATE Compute prediction result $\hat{G}_{i + 1}$ using Eq. \eqref{13}.
            \STATE Refine prediction result using Eq. \eqref{14}.
            \STATE Compute loss using Eq. \eqref{19}.
        \ENDFOR
        \STATE Backpropagate and update model parameters.
    \ENDFOR
\ENDFOR
\STATE Return optimized parameter $\Theta_*$.
\end{algorithmic}
\end{algorithm}

\vspace{-10pt}
\subsection{Model Complexity Analysis}
For a graph snapshot $G_t$, the micro feature extraction sub-module utilizes a weighted GAT to aggregate node attributes. The time complexity of the weighted GAT remains consistent with that of the original GAT \cite{velickovic2017graph}, which is $O(|V|DD_X + |E^t|D_X)$. The meso feature extraction sub-module consists of  two primary operations: community detection and meso-scale pooling. Community detection is achieved using the Louvain algorithm \cite{blondel2008fast}, which has a time complexity of $O(|V| \log |V|)$.
In meso-scale pooling, for a community $G_{t,c}$, calculating the sum of link weights for nodes has a time complexity of $O(E_{t,c})$, while the weighted pooling of node features adds further time complexity of $O(|V_{t,c}|D)$. Assuming that $G_t$ contains $M$ communities, each with an average of $|V_{t,a}|$ nodes and $|E_{t,a}|$ links, the total time complexity of the meso-scale pooling is $O(M(|E_{t,a}| + |V_{t,a}|D))$.
The macro feature extraction sub-module performs a similar pooling operation over the entire graph, resulting in a time complexity of $O(|E^t| + |V|D)$. For the temporal feature extraction module, the time complexity depends on the length of the input sequence and the dimensionality of the hidden state, yielding a complexity of $O(l|V|D_L(3D + D_L))$.
Finally, the feature decoding module utilizes a neural network consisting of two fully connected layers. For each sample, the time complexity of a single  layer is $O(|V|^2D)$.

\section{Experiment}\label{Experiment}
In this section, we first introduce the datasets used in the experimental evaluation. Next, we present the two performance metrics employed to assess the prediction methods. We then describe the baseline methods selected for comparison with our model, MUST. Finally, we present the experimental results to validate the performance of MUST. We will make the code and datasets public at \href{https://github.com/wufangr/MUST}{https://github.com/wufangr/MUST}. 

\begin{table*}[!t]
\renewcommand{\arraystretch}{1.2}
\caption{UANET simulation parameter settings}
\label{tab1}
\centering
\begin{tabularx}{\textwidth}{>{\centering\arraybackslash}X>{\centering\arraybackslash}X>{\centering\arraybackslash}X>{\centering\arraybackslash}X>{\centering\arraybackslash}X>{\centering\arraybackslash}X>{\centering\arraybackslash}X}
\hline
\makecell{Number of\\ UAVs} & \makecell{Simulation area\\ size ($\text{km}^\text{2}$)}
 & \makecell{Flight\\ speed (m/s)} & \makecell{Communication\\ radius (m)}  & \makecell{Simulation\\ duration (s)} & \makecell{Sampling\\ interval (s)}  \\ 
\hline
100            & 100                         & 25$\sim$35          & 1000$\sim$2000                       & 800                     & 10                      \\
\hline
\end{tabularx}
\vspace{-10pt}
\end{table*}

\begin{table}[!t]
\vspace{-10pt}
\renewcommand{\arraystretch}{1.2}
\caption{The statistics of UANET snapshot sequences}
\label{tab2}
\centering
\begin{tabularx}{0.48\textwidth}{c|>{\centering\arraybackslash}X>{\centering\arraybackslash}X>{\centering\arraybackslash}X>{\centering\arraybackslash}X}
\hline
Dataset & Min. edges & Max. edges & Avg. edges & Avg. density  \\ 
\hline
RW             & 377        & 451        & 416.4      & 0.0843             \\
MG             & 368        & 459        & 412.1      & 0.0833             \\
RPG            & 437        & 697        & 559.7      & 0.1167             \\
GM             & 510        & 589        & 546.2      & 0.1105             \\
\hline
\end{tabularx}
\end{table}

\vspace{-10pt}
\subsection{Datasets}
Due to the scarcity of realistic UANET datasets, we generate UAV mobility data using four representative UAV mobility models. The simulation parameters for these models are presented in Table \ref{tab1}. The four mobility models are outlined below.
\begin{itemize}
\item{\textbf{The Random Walk (RW) mobility model} \cite{chiang20042}: This model simulates unpredictable movements commonly observed in natural entities. At regular intervals, each UAV selects a random direction and speed, maintaining this trajectory for a specific duration before updating its direction and speed. The RW model is particularly useful for area coverage tasks.}
\item{\textbf{The Manhattan Grid (MG) mobility model} \cite{kalyanam2020graph}: This model simulates UAV movement within a grid-like urban road structure. At intersections, UAVs can turn left, right, or continue straight, based on predefined probabilities. The MG model is ideal for traffic monitoring and urban surveillance tasks.}
\item{\textbf{The Reference Point Group (RPG) mobility model} \cite{hong1999group}: In this group mobility model, UAVs follow a common reference point instead of moving independently. The reference point periodically selects new destinations and moves at a constant speed. RPG is well-suited for agricultural management.}
\item{\textbf{The Gauss-Markov (GM) mobility model } \cite{camp2002survey}: This model adjusts UAV movement based on past speed and direction. Each UAV starts with an initial speed and direction, which are then updated over time using predefined rules to ensure smooth transitions. GM is ideal for coverage tasks.}
\end{itemize}

These mobility models generate  diverse mobility data. Using Eq. (1), we derive the UAV topology at each time step. Data collection begins once the number of links stabilizes, and we sample the UAV data every 10 seconds, resulting in a total of 80 snapshots per mobility model. The detailed statistics of the graph snapshot sequences are provided in  Table \ref{tab2}, which highlights the sparsity of the snapshots.

\vspace{-10pt}
\subsection{Evaluation Metrics}
To evaluate the performance of our model in UANET link prediction, we use two key metrics: the Area Under the ROC Curve (AUC) and the Area Under the Precision-Recall Curve (AUPRC).
\begin{itemize}
\item{\textbf{AUC:} The ROC curve illustrates the relationship between the True Positive Rate (TPR) and the False Positive Rate (FPR) at various threshold levels. AUC is widely used in dynamic link prediction tasks and assesses the model's performance across different classification thresholds, offering a comprehensive measure of overall effectiveness.}
\item{\textbf{AUPRC:} The PR curve shows the trade-off between Precision and Recall at various thresholds. Unlike AUC, AUPRC focuses more on detecting the positive class. In sparse UANETs, where non-existent links significantly outnumber existing ones, AUPRC provides a more accurate reflection of the model’s ability to predict existing links in imbalanced datasets.}
\end{itemize}

\vspace{-10pt}
\subsection{Baseline Methods}
To evaluate the performance of our method, we compare it with several popular and state-of-the-art methods. The following five baseline methods were selected for comparison:
\begin{itemize}
	\item{\textbf{Node2Vec}\cite{grover2016node2vec}:  Node2Vec is a well-known static graph embedding method that generates node sequences through biased random walks and learns node vector representations using the Skip-gram model. Since Node2Vec is designed for static networks, we adapt it to our task by aggregating the first $l-1$ snapshots in each sample, applying Node2Vec to the aggregated graph, and then predicting $G_l$.}
	\item{\textbf{DySAT}\cite{sankar2020dysat}: DySAT is a dynamic graph embedding method that leverages self-attention mechanisms. It incorporates both structural and temporal self-attention layers to capture structural features and capture temporal evolution patterns, respectively. Specifically, structural self-attention is applied to each snapshot in the historical sequence to obtain static node representations, while temporal self-attention captures the evolution of these representations, resulting in dynamic graph node embeddings.}
	\item{\textbf{EvolveGCN}\cite{pareja2020evolvegcn}: EvolveGCN evolves the parameters of a GCN to learn the representation of dynamic graphs. At each time step, a GCN acts as a feature extractor, while a RNN adjusts the GCN's parameters. In this approach, only the RNN is trained, and the GCN parameters are computed by the RNN.  EvolveGCN  comes in two versions: EvolveGCN-O and EvolveGCN-H, which differ in the type of RNN used and the inputs provided to the RNN.}
	\item{\textbf{E-LSTM-D}\cite{chen2019lstm}: E-LSTM-D is an end-to-end deep learning model designed  for dynamic network link prediction. It integrates stacked LSTMs into an encoder-decoder framework to learn low-dimensional network representations, extract non-linear features, and model the temporal evolution of the network.  This approach enables the reconstruction of the full graph structure.  To address network sparsity, E-LSTM-D incorporates improvements during the training process.}
	\item{\textbf{Top-Sequential-Stacking}\cite{he2024sequential}: Top-Sequential-Stacking is a framework for link prediction that sequentially stacks topological features. At each time step, static topological features are computed from sampled node pairs and combined into a feature vector, which is then used for dynamic link prediction. This method performs well in both partially observed and fully unobserved settings.}   
\end{itemize}

\begin{table*}[!t]
\renewcommand{\arraystretch}{1.2}
\caption{Hyperparameter settings of MUST}
\label{tab3}
\centering
\begin{tabularx}{\textwidth}{>{\centering\arraybackslash}p{2.4cm} >{\centering\arraybackslash}p{2.4cm} >{\centering\arraybackslash}p{2.4cm}|>{\centering\arraybackslash}X>{\centering\arraybackslash}X>{\centering\arraybackslash}X>{\centering\arraybackslash}X>{\centering\arraybackslash}X}
\hline
\multicolumn{3}{c|}{Network dimension configurations}   & \multicolumn{5}{c}{Hyperparameters}                           \\ 
\hline
GAT             & LSTM              & FC                & $r$     & $\varepsilon$ & $\beta$ & $\eta$ & $\lambda$  \\
\hline
{[}132, 100, 44] & {[}132, 256, 256] & {[}256, 256, 100] & 0.0005 & 10         & 0.001   & 0.5    & 0.0005    \\
\hline
\end{tabularx}
\end{table*}

\begin{table*}[!t]
\vspace{-10pt}
\renewcommand{\arraystretch}{1.2}
\caption{AUC and AUPRC for All the Methods Across the Four Mobility Models}
\label{tab4}
\centering
\begin{tabularx}{\textwidth}{c|>{\centering\arraybackslash}X>{\centering\arraybackslash}X>{\centering\arraybackslash}X>{\centering\arraybackslash}X>{\centering\arraybackslash}X>{\centering\arraybackslash}X>{\centering\arraybackslash}X>{\centering\arraybackslash}X}
\hline
\multirow{2}{*}{Methods} & \multicolumn{2}{c}{RW}            & \multicolumn{2}{c}{MG}            & \multicolumn{2}{c}{RPG}           & \multicolumn{2}{c}{GM}             \\ 
\cline{2-9}
                         & AUC             & AUPRC           & AUC             & AUPRC           & AUC             & AUPRC           & AUC             & AUPRC            \\ 
\hline
Node2Vec                 & 0.5321          & 0.0902          & 0.5292          & 0.0898          & 0.5124          & 0.1112          & 0.5259          & 0.1155           \\
DySAT                    & 0.6302          & 0.1370          & 0.6234          & 0.1304          & 0.6921          & 0.2020          & 0.6171          & 0.1835           \\
EvolvegGCN-O             & 0.8187          & 0.3435          & 0.7988          & 0.3181          & 0.8020          & 0.3427          & 0.8088          & 0.3548           \\
EvolvegGCN-H             & 0.8334          & 0.3332          & 0.7939          & 0.2986          & 0.8101          & 0.3365          & 0.7864          & 0.3340           \\
E-LSTM-D                 & 0.8850          & 0.3984          & 0.8537          & 0.3184          & 0.6380          & 0.1829          & 0.8046          & 0.3245           \\
Top-Sequential-Stacking  & 0.8235          & 0.2240          & 0.7862          & 0.2059          & 0.5245          & 0.1259          & 0.7142          & 0.2075           \\
MUST                    & \textbf{0.9367} & \textbf{0.5571} & \textbf{0.9268} & \textbf{0.5119} & \textbf{0.8272} & \textbf{0.3917} & \textbf{0.8954} & \textbf{0.4927}  \\
\hline
\end{tabularx}
\end{table*}

\vspace{-10pt}
\subsection{Parameter settings}
We set the window size \( l = 11 \). By sliding this window over the sequence of 80 snapshots with a step size of 1, we generate 70 samples.  The first 50 samples are used for training to learn the model parameters, while the remaining 20 samples are reserved for testing and evaluating model performance.  To ensure a fair comparison, the first 10 snapshots in each sample are designated as historical snapshots, and the last snapshot is used as the prediction target across all methods. 

For Node2Vec, DySAT, and EvolveGCN, the embedding dimension is set to 132. Similarly, for E-LSTM-D, the node representation dimension prior to being fed into the decoder is also set to 132. For Top-Sequential-Stacking, the node feature dimension is determined by the number of topological features used, as described in the original paper \cite{he2024sequential}. The remaining parameters for the baseline methods are configured according to their  original papers \cite{he2024sequential,pareja2020evolvegcn,chen2019lstm,grover2016node2vec,sankar2020dysat}. 

In MUST, the dimensions for micro-scale, meso-scale, and macro-scale representations are set to 44 each, ensuring a total node feature dimension of 132.
The multi-scale feature extraction module employs a 2-layer weighted GAT, the temporal feature extraction module utilizes a 2-layer stacked LSTM, and the feature decoding module incorporates a 2-layer fully connected neural network. Detailed parameter configurations for these layers are provided in Table \ref{tab3}, where the numbers in square brackets denote the input, intermediate, and output dimensions for each network layer. Additionally, \( r \) represents the learning rate, and the   hyperparameters  $\varepsilon$, $\beta$, $\eta$ and $\lambda$  are used to regulate the terms in the loss function.

\vspace{-10pt}
\subsection{Experimental Results}

\begin{figure*}[!t]
\includegraphics[width=1\textwidth]{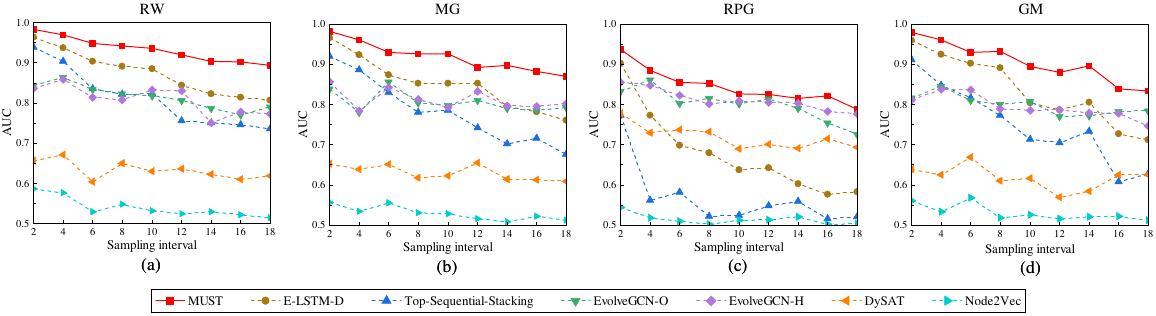}
\caption{AUC vs. Sampling interval for all methods across the four mobility models.   MUST consistently outperforms the baseline methods across all sampling intervals.}
\label{fig3}
\end{figure*}

\begin{figure*}[!t]
\vspace{-10pt}
\includegraphics[width=1\textwidth]{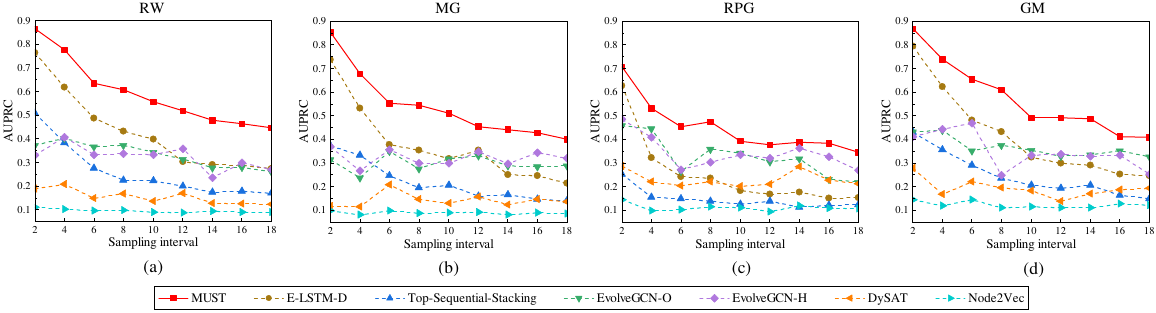}
\caption{AUPRC    vs. Sampling interval for all methods across the four mobility models.   MUST consistently outperforms the baseline methods across all sampling intervals.}
\label{fig4}
\vspace{-10pt}
\end{figure*}


\subsubsection{Comparison with Baselines}
Table \ref{tab4} presents the AUC and AUPRC results for MUST and the baseline methods across the four UANET datasets. The results reveal that MUST consistently outperforms all   the other methods across every dataset. Specifically, MUST improves AUC by 2.1\% to 10.7\% and AUPRC by 14.3\% to 60.9\% compared to the second-best method. The improvement in AUC highlights MUST's superior overall learning and prediction capabilities. The significant increase in AUPRC underscores its effectiveness in predicting links within sparse UANETs, demonstrating the robustness of the proposed loss function.
Among the baseline methods, Node2Vec performs the worst, with results approximating random predictions. This is attributed to its static network embedding approach, which fails to capture the dynamic and temporal nature of UANETs effectively. DySAT, incorporating temporal information, performs better than Node2Vec. However, its performance is hindered by the separation of embedding learning and link prediction, as it lacks an end-to-end training mechanism.  EvolveGCN, E-LSTM-D, and Top-Sequential-Stacking achieve AUC values approaching those of MUST in several cases, but their AUPRC scores remain significantly lower, highlighting their limitations in handling UANET sparsity. Notably, Top-Sequential-Stacking performs particularly poorly on the RPG mobility model, showcasing the instability of methods that rely heavily on traditional topological features. In contrast, MUST effectively addresses the challenges posed by the high dynamics and sparsity of UANETs, delivering consistently stable and superior performance across all four mobility models. This highlights its robustness and adaptability in dynamic network environments.

\subsubsection{Impact of Snapshot Sampling Interval}
In the earlier experiments, the UAV datasets are sampled at fixed 10-second intervals. However, the sampling interval significantly affects  model performance, as larger intervals weaken the temporal dependencies between snapshots. To investigate this, we evaluate sampling intervals ranging from 2 to 18 seconds, increasing in 2-second increments. To maintain consistency across simulations, the duration is adjusted between 160 and 1440 seconds, generating 80 snapshots per mobility model. Other simulation parameters in Table \ref{tab1} remain unchanged. Through simulation, we generate a total of nine snapshot sequences with different sampling intervals for each mobility model.
\begin{figure*}[!t]
\includegraphics[width=1\textwidth]{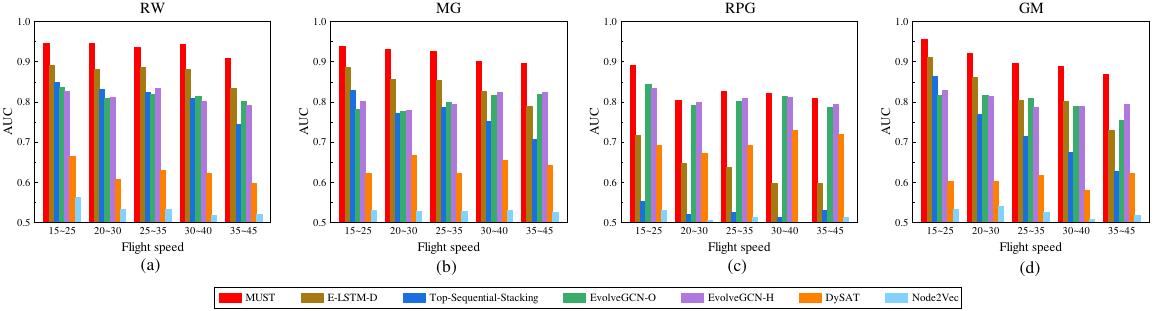}
\caption{AUC vs. Flight speed for all methods across the four mobility models. MUST consistently outperforms the baseline methods across all flight speeds.}
\label{fig5}
\end{figure*}

\begin{figure*}[!t]
\includegraphics[width=1\textwidth]{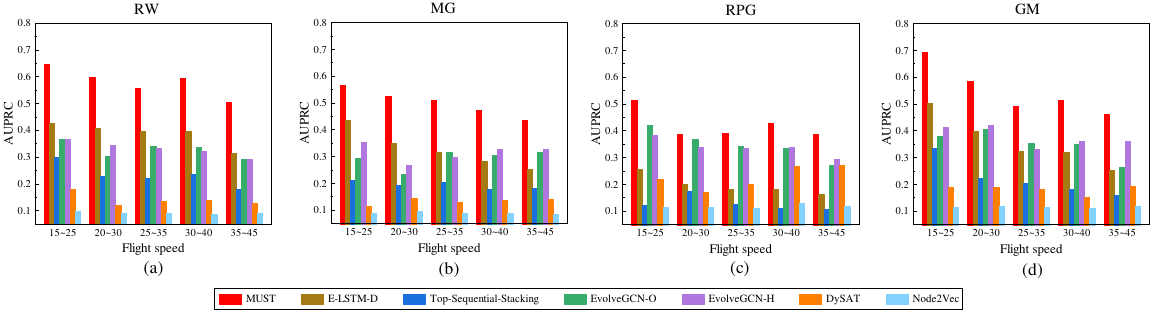}
\caption{AUPRC vs. Flight speed for all methods across the four mobility models. MUST consistently outperforms the baseline methods across all flight speeds.}
\label{fig6}
\vspace{-10pt}
\end{figure*}
Figs. \ref{fig3} and \ref{fig4} display the AUC and AUPRC trends, respectively, for the four mobility models across various sampling intervals. The results reveal a consistent decline in both AUC and AUPRC for MUST, E-LSTM-D, and Top-Sequential-Stacking as the sampling interval increases.  EvolveGCN exhibits a similar trend in most scenarios.  The performance degradation  reinforces that  smaller intervals allow models to better capture temporal dependencies, while larger intervals result in significant information loss. In addition, DySAT and Node2Vec show no consistent performance trends with increasing sampling intervals, reflecting their limited capability to capture temporal dynamics effectively. 
Despite the overall negative impact of increasing sampling intervals on link prediction performance, MUST consistently outperforms all baseline methods across mobility models and sampling intervals. Notably, MUST achieves significant AUPRC improvements in the RW, MG, and GM mobility models, further validating its ability to handle network sparsity   in link prediction.

\subsubsection{Impact of Flight Speed}
In prior experiments, UAV flight speeds are fixed between 25 m/s and 35 m/s.  However, flight speed can significantly influence prediction performance, as higher speeds cause more frequent changes in network topology. To evaluate this impact, we tested five different flight speed ranges: 15$\sim$25 m/s, 20$\sim$30 m/s, 25$\sim$35 m/s, 30$\sim$40 m/s, and 35$\sim$45 m/s, while keeping other simulation parameters constant. This setup produces five distinct datasets for each mobility model, corresponding to the different speed ranges.


Figs. 5 and 6 depict the AUC and AUPRC trends across the four mobility models for each method under varying flight speeds. The results show that, with the increase in flight speed, the link prediction performance tends to decrease for MUST, E-LSTM-D, and Top-Sequential-Stacking, while the performance fluctuates for the other methods. The performance of all methods in the RPG model is lower than in the other mobility models.
As flight speed increases, the network topology changes more frequently, reducing the dependency between network snapshots and leading to a decrease in link prediction performance. The insensitivity of some methods, such as Node2Vec, to flight speed can be attributed to their lack of reliance on temporal dynamics.
Overall, MUST consistently outperforms all baseline methods across mobility models and flight speeds, highlighting its capability to handle dynamic and rapidly changing network topologies.

\begin{table*}[htb]
\renewcommand{\arraystretch}{1.2}
\caption{AUC and AUPRC for MUST Across the Four Mobility Models: Only Micro-Scale Features,  No Micro-Scale Features, No Meso-Scale Features, No Macro-Scale Features, and Multi-Scale Features.}
\label{tab5}
\centering
\begin{tabularx}{\textwidth}{c|>{\centering\arraybackslash}X>{\centering\arraybackslash}X>{\centering\arraybackslash}X>{\centering\arraybackslash}X>{\centering\arraybackslash}X>{\centering\arraybackslash}X>{\centering\arraybackslash}X>{\centering\arraybackslash}X}
\hline
\multirow{2}{*}{Models} & \multicolumn{2}{c}{RW}            & \multicolumn{2}{c}{MG}   & \multicolumn{2}{c}{RPG}           & \multicolumn{2}{c}{GM}             \\ 
\cline{2-9}
                        & AUC             & AUPRC           & AUC             & AUPRC  & AUC             & AUPRC           & AUC             & AUPRC            \\ 
\hline
Only-micro                & 0.9356          & 0.5565          & 0.9181          & 0.5056 & 0.8237          & 0.3751          & 0.8902          & 0.4913           \\
No-micro                & 0.9168          & 0.4855          & 0.9114          & 0.4464 & 0.7931          & 0.3281          & 0.8690          & 0.4298           \\
No-meso                 & 0.9366          & 0.5585          & 0.9225          & 0.5104 & \textbf{0.8282}          & 0.3878          & 0.8929          & 0.4864  \\
No-macro                & 0.9262          & \textbf{0.5627}          & 0.9245 & 0.4993 & 0.8251          & 0.3872          & 0.8949          & 0.4897           \\
Multi-Scale             & \textbf{0.9367} & 0.5571 & \textbf{0.9268}          & \textbf{0.5119} & 0.8272 & \textbf{0.3917} & \textbf{0.8954} & \textbf{0.4927}  \\
\hline
\end{tabularx}
\vspace{-10pt}
\end{table*}
\vspace{2pt}
\subsubsection{Effectiveness of Multi-Scale Feature Extraction}
To evaluate the effectiveness of multi-scale feature extraction module in UANET link prediction, we conduct an ablation study. This module extracts features at three levels: micro-scale, meso-scale, and macro-scale, and combines them through feature concatenation to generate the final multi-scale representation. In the ablation study, we compare the performance of the complete multi-scale features with variations: 1) Only micro-scale features and 2) features from one specific scale are excluded.

The results, summarized in Table \ref{tab5}, clearly show that using the complete multi-scale features provides the best link prediction performance in most cases. Using only micro features or removing features from any single scale causes performance degradation.   Among the scales, micro-scale features prove to be the most critical. Excluding micro-scale features led to the largest drop in performance compared to excluding meso-scale or macro-scale features. This highlights the pivotal role of micro-scale features in capturing fine-grained, localized patterns that are essential for effective link prediction in highly dynamic UANET environments. The findings confirm that integrating multi-scale features enhances the model’s ability to capture complex structural and temporal dependencies, significantly improving its predictive accuracy.

\section{Conclusion}\label{Conclusion}
We propose a novel link prediction model, MUST, designed specifically to address the challenges of high dynamics and sparse topology in UANETs. MUST leverages multi-scale feature extraction to capture UANET structural features across micro-, meso-, and macro-scales. Additionally, it integrates temporal feature extraction using a stacked LSTM to model the temporal dependencies between snapshots effectively. To enhance its performance, we design a tailored loss function that helps the model capture evolving patterns and mitigate the impact of UANET sparsity. MUST is evaluated on diverse UAV datasets generated from four representative mobility models, benchmarking its performance in terms of AUC and AUPRC against five state-of-the-art methods. The experimental results demonstrate that MUST consistently outperforms baseline models across various sampling intervals and flight speeds. Furthermore, an ablation study validates the effectiveness of the multi-scale feature extraction mechanism in improving link prediction accuracy. Notably, our work is particularly useful in adversarial environments, where the task is to predict the topologies of rival UANETs. Moreover, the proposed model can be extended to broader applications in dynamic network prediction tasks.




\begin{IEEEbiography}[{\includegraphics[width=1in,height=1.25in,clip,keepaspectratio]{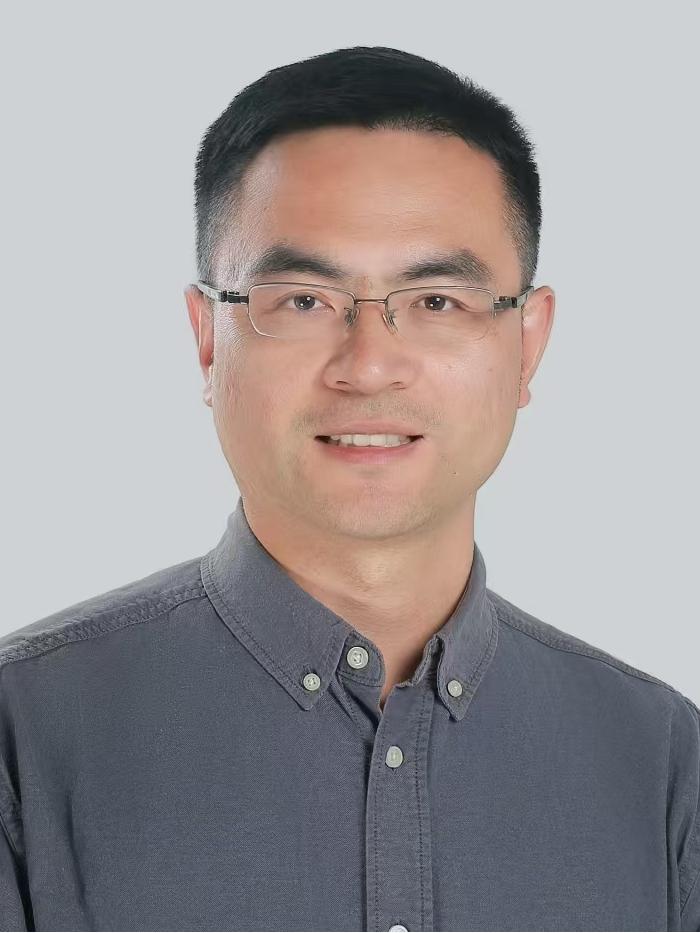}}]{Cunlai Pu}
received the Ph.D. degree in information and communication engineering from Southeast University, Nanjing, China, in 2012. He is currently an Associate Professor with the School of Computer Science and Engineering, Nanjing University of Science and Technology, Nanjing, China. His interests include network science, communication systems, and network optimization.
\end{IEEEbiography}

\vspace{-20pt}
\begin{IEEEbiography}[{\includegraphics[width=1in,height=1.25in,clip,keepaspectratio]{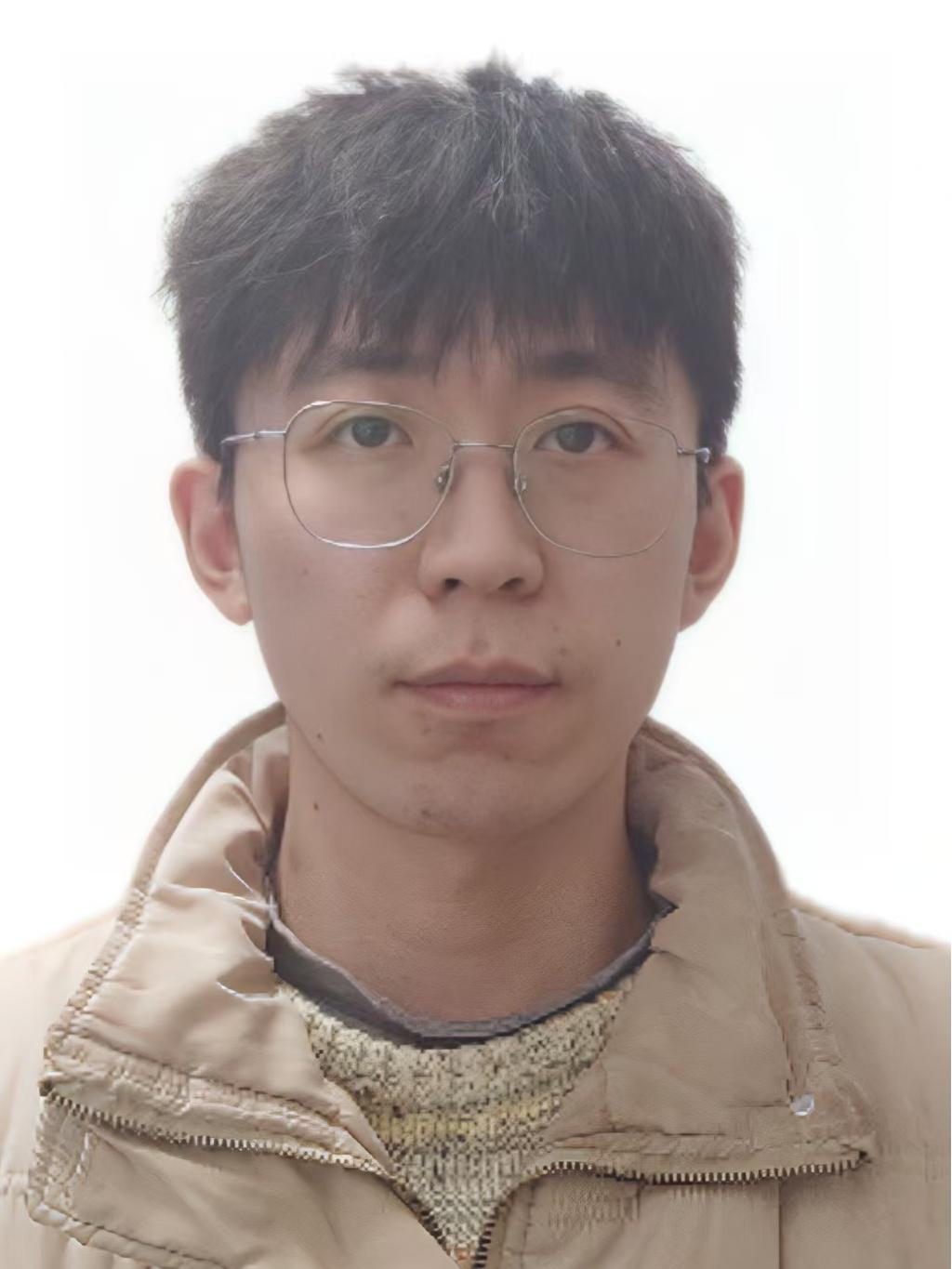}}]{Fangrui Wu}
received the B.S. degree from Guangdong University of Technology, Guangzhou, China, in 2023. He is currently working toward the M.S. degree in computer science with the School of Computer Science and Engineering, Nanjing University of Science and Technology. His research interests include dynamic link prediction and graph machine learning.
\end{IEEEbiography}

\vspace{-20pt}
\begin{IEEEbiography}[{\includegraphics[width=1in,height=1.25in,clip,keepaspectratio]{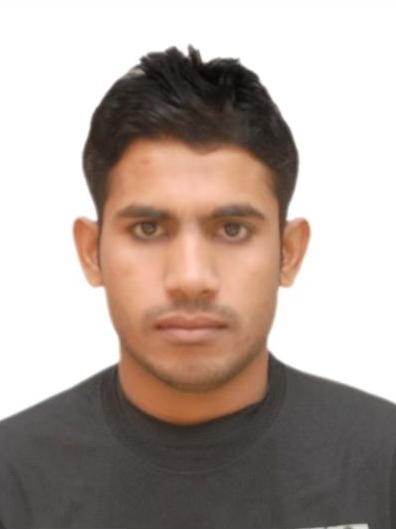}}]{Rajput Ramiz Sharafat}
	 received his B.S. (Hons) degree in Information Technology from Sindh Agriculture University, Sindh, Pakistan. He then received his M.S. degree in Computer Science and Technology from the Nanjing University of Science and Technology, Nanjing, China, where he was awarded the CSC Scholarship and also received the Best Scientific Publication Award from the School of International Education, NJUST and the Embassy of Pakistan (Shanghai Consulate). He is currently pursuing a Ph.D. degree at the University of Science and Technology of China, Hefei, China. His research interests include content caching, wireless communication, vehicular networks, federated learning,  and deep Reinforcement Learning. 
\end{IEEEbiography}

\vspace{-20pt}
\begin{IEEEbiography}[{\includegraphics[width=1in,height=1.25in,clip,keepaspectratio]{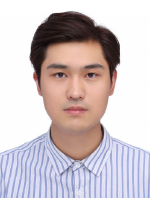}}]{Guangzhao Dai}
received the B.S. degree in automation from Wuxi Taihu University, China, in 2019. He is currently pursuing the Ph.D. degree with the School of Computer Science and Engineering, Nanjing University of Science and Technology. His research interests include video understanding, video action recognition, and egocentric action recognition.
\end{IEEEbiography}

\vspace{-20pt}
\begin{IEEEbiography}[{\includegraphics[width=1in,height=1.25in,clip,keepaspectratio]{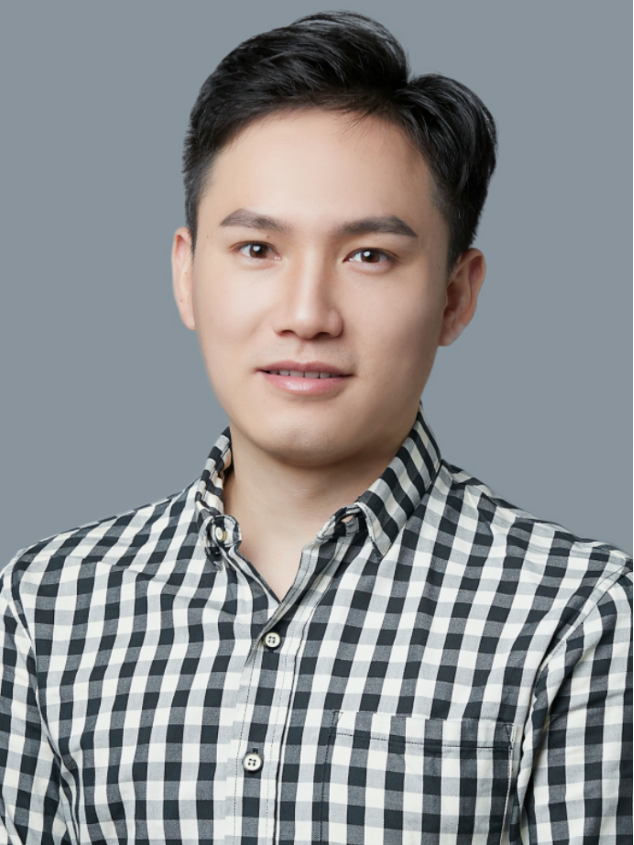}}]{Xiangbo Shu}
(Senior Member, IEEE) received the Ph.D. degree from  Nanjing University of Science and Technology in 2016. From 2014 to 2015, he was a Visiting Scholar with  National University of Singapore, Singapore. He is currently a Professor with the School of Computer Science and Engineering, Nanjing University of Science and Technology, China. His current research interests include computer vision, multimedia and machine learning. He has authored over 100 journals and conference papers in these areas, including IEEE TPAMI, TNNLS, TIP; CVPR, ICCV, ECCV, and ACM MM, etc. He has served as the editorial boards of the IEEE TNNLS, and the IEEE TCSVT. He is also the Member of ACM, the Senior Member of CCF, and the Senior Member of IEEE.
\end{IEEEbiography}

\vfill

\end{document}